\documentclass[letterpaper, 10 pt, conference]{ieeeconf}  

\IEEEoverridecommandlockouts                              

\usepackage{amsthm,amsfonts}
\usepackage{graphicx}
\usepackage{xcolor}
\usepackage[maxbibnames=99,backend=biber]{biblatex} 
\usepackage{tabularx}
\usepackage{float}
\usepackage{hyperref}
\usepackage{caption}
\usepackage{graphicx}
\usepackage{subcaption}
\usepackage[utf8]{inputenc}
\usepackage[T1]{fontenc}
\usepackage{textcomp}
\usepackage{tablefootnote}
\usepackage{amsmath,amssymb}
\let\labelindent\relax
\usepackage[inline]{enumitem}
\usepackage{threeparttable}
\usepackage{siunitx}
\usepackage{algpseudocode}
\usepackage{algorithm}
\usepackage{pythonhighlight}
\usepackage{tcolorbox}
\tcbuselibrary{breakable,xparse,skins}

\usepackage{svg}

\definecolor{bg}{gray}{0.95}

\title{\LARGE \bf
Fine-Grained Open-Vocabulary Object Detection with Fined-Grained Prompts: Task, Dataset and Benchmark
}

\author{Ying Liu$^{1}$, Yijing Hua$^{1}$, Haojiang Chai$^{1}$, Yanbo Wang$^{1}$, TengQi Ye*
\thanks{$^{1}$Ying Liu, Yijing Hua, Haojiang Chai and Yanbo Wang are with the department of software engineering,
        Northeastern University, China.}%
\thanks{*TengQi Ye ({\tt\small yetengqi@gmail.com}) is the corresponding author. His work was done in Articul8 AI.}
}

\begin{document}

\maketitle
\thispagestyle{empty}
\pagestyle{empty}

\begin{abstract}
Open-vocabulary detectors are proposed to locate and recognize objects in novel classes. However, variations in vision-aware language vocabulary data used for open-vocabulary learning can lead to unfair and unreliable evaluations. Recent evaluation methods have attempted to address this issue by incorporating object properties or adding locations and characteristics to the captions. Nevertheless, since these properties and locations depend on the specific details of the images instead of classes, detectors can not make accurate predictions without precise descriptions provided through human annotation.
This paper introduces 3F-OVD, a novel task that extends supervised fine-grained object detection to the open-vocabulary setting. Our task is intuitive and challenging, requiring a deep understanding of \textit{F}ine-grained captions and careful attention to \textit{F}ine-grained details in images in order to accurately detect \textit{F}ine-grained objects. Additionally, due to the scarcity of qualified fine-grained object detection datasets, we have created a new dataset, NEU-171K, tailored for both supervised and open-vocabulary settings. We benchmark state-of-the-art object detectors on our dataset for both settings. Furthermore, we propose a simple yet effective post-processing technique. Our data, annotations and codes are available at \url{https://github.com/tengerye/3FOVD}.

\end{abstract}

\section{INTRODUCTION}
Despite its success, supervised object detection relies on human annotations for bounding boxes and corresponding class labels, which are usually expensive and time-consuming. Once trained, these detectors can only recognize objects from a predefined set of classes, making them incapable of identifying novel objects. 
Open-vocabulary detectors \cite{gu2021open,zareian2021open} were proposed to extend detectors to generalize from annotated base classes to novel classes. During inference, these detectors can use captions that contains the visual descriptions of desired unseen classes to predict bounding boxes \cite{wu2024towards}.

Open-vocabulary learning, which leverages visual-related language vocabulary data, is essential for open-vocabulary detectors. However, different detectors can utilize distinct datasets for pre-training, leading to inconsistencies in evaluation. For example, ViLD \cite{gu2021open} mainly uses LVIS \cite{gupta2019lvis} for pre-training while Gounding DINO \cite{liu2023grounding} utilizes up to six datasets for pre-training. Consequently, as noted in \cite{wang2024ov}, open-vocabulary algorithms face a significant risk of data leakage, where novel or closely related classes may inadvertently be included in pre-training data.

\begin{figure}[!h]
     \centering
     \begin{subfigure}[b]{0.23\textwidth}
         \centering
         \includegraphics[width=\textwidth]{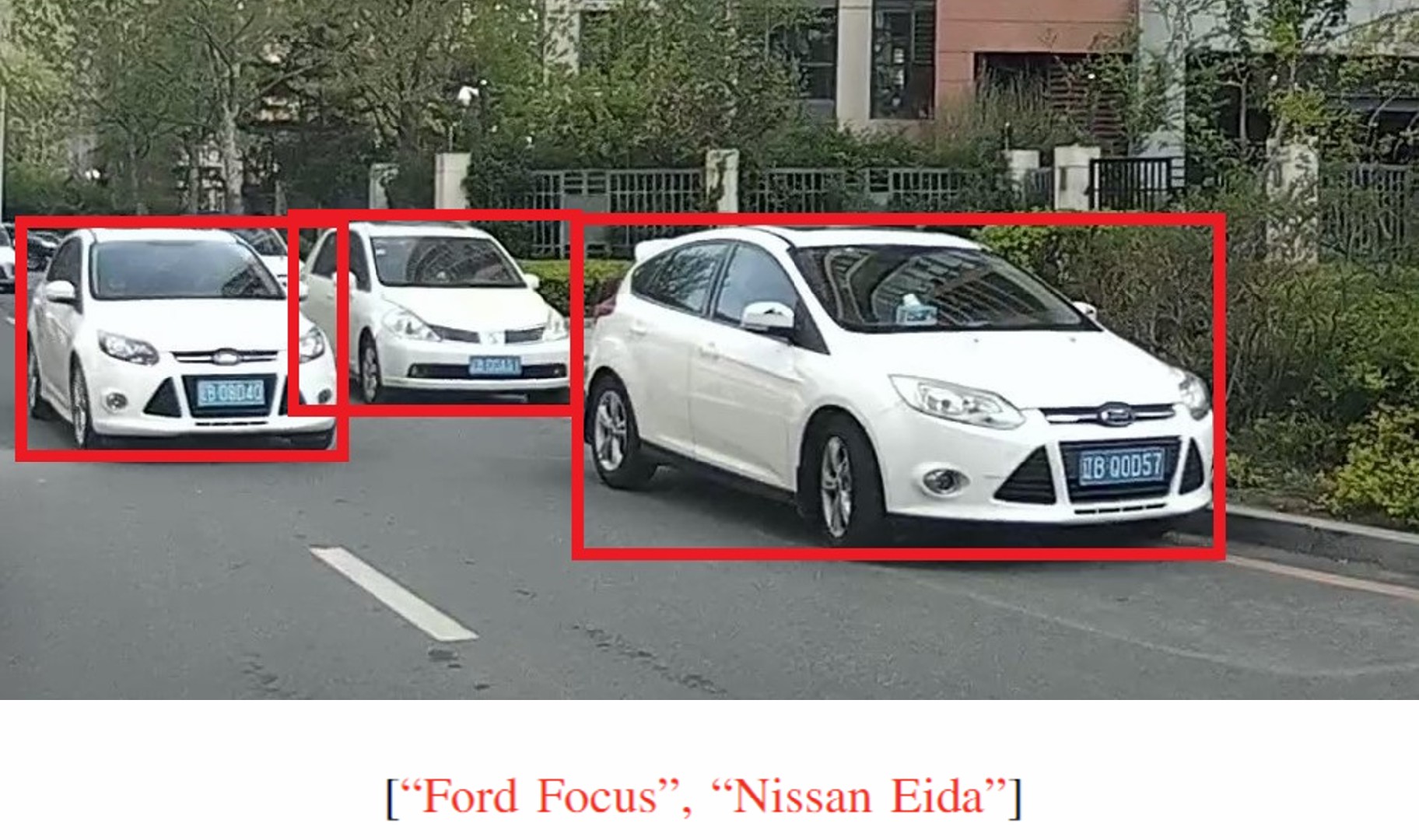}
         \caption{Supervised FG-OD.}
     \end{subfigure}
     \begin{subfigure}[b]{0.23\textwidth}
         \centering
             \includegraphics[width=\textwidth]{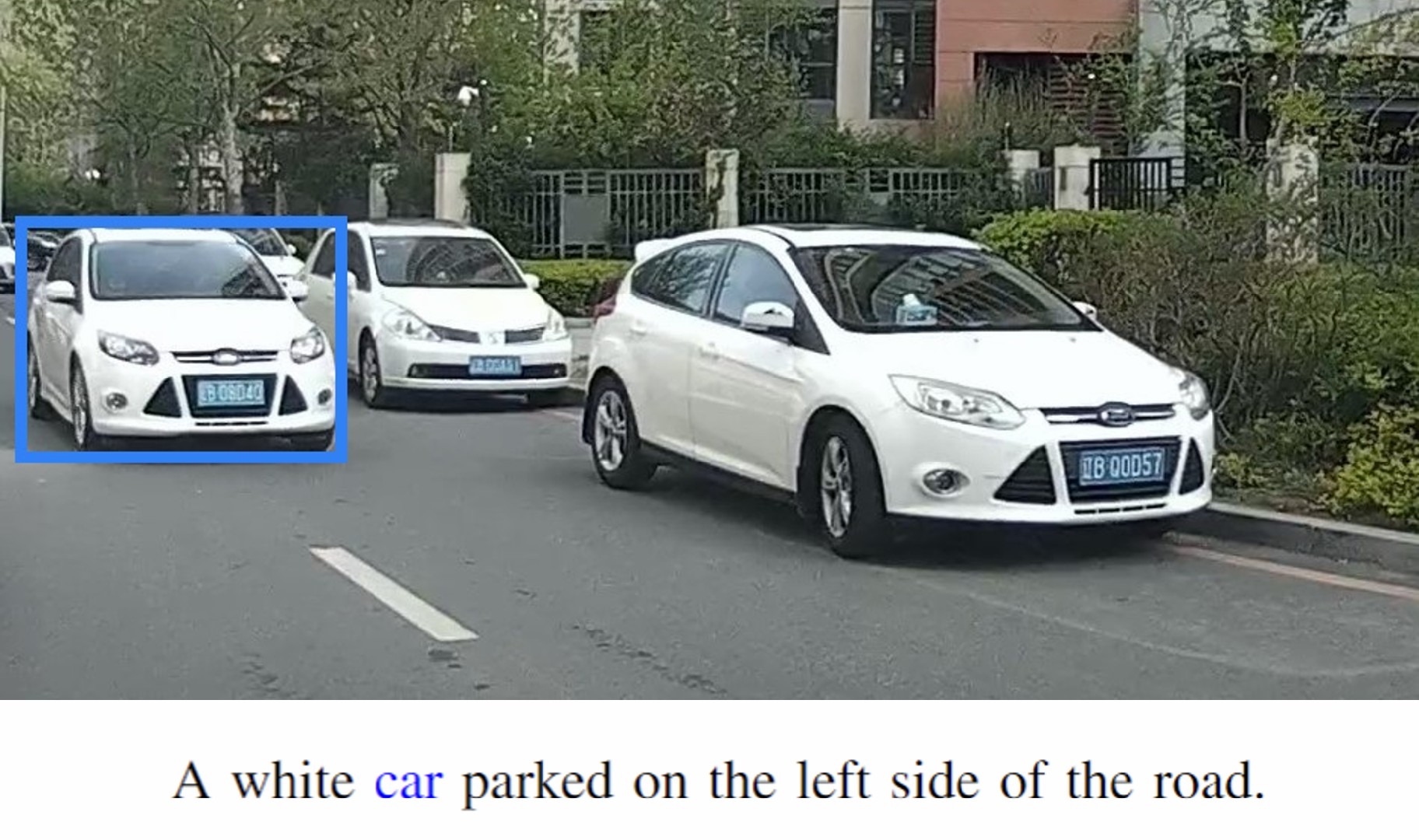}
         \caption{OV-VG\cite{wang2024ov}.}
         \label{fig:ov-vg}
     \end{subfigure}
     \begin{subfigure}[b]{0.23\textwidth}
         \centering
         \includegraphics[width=\textwidth]{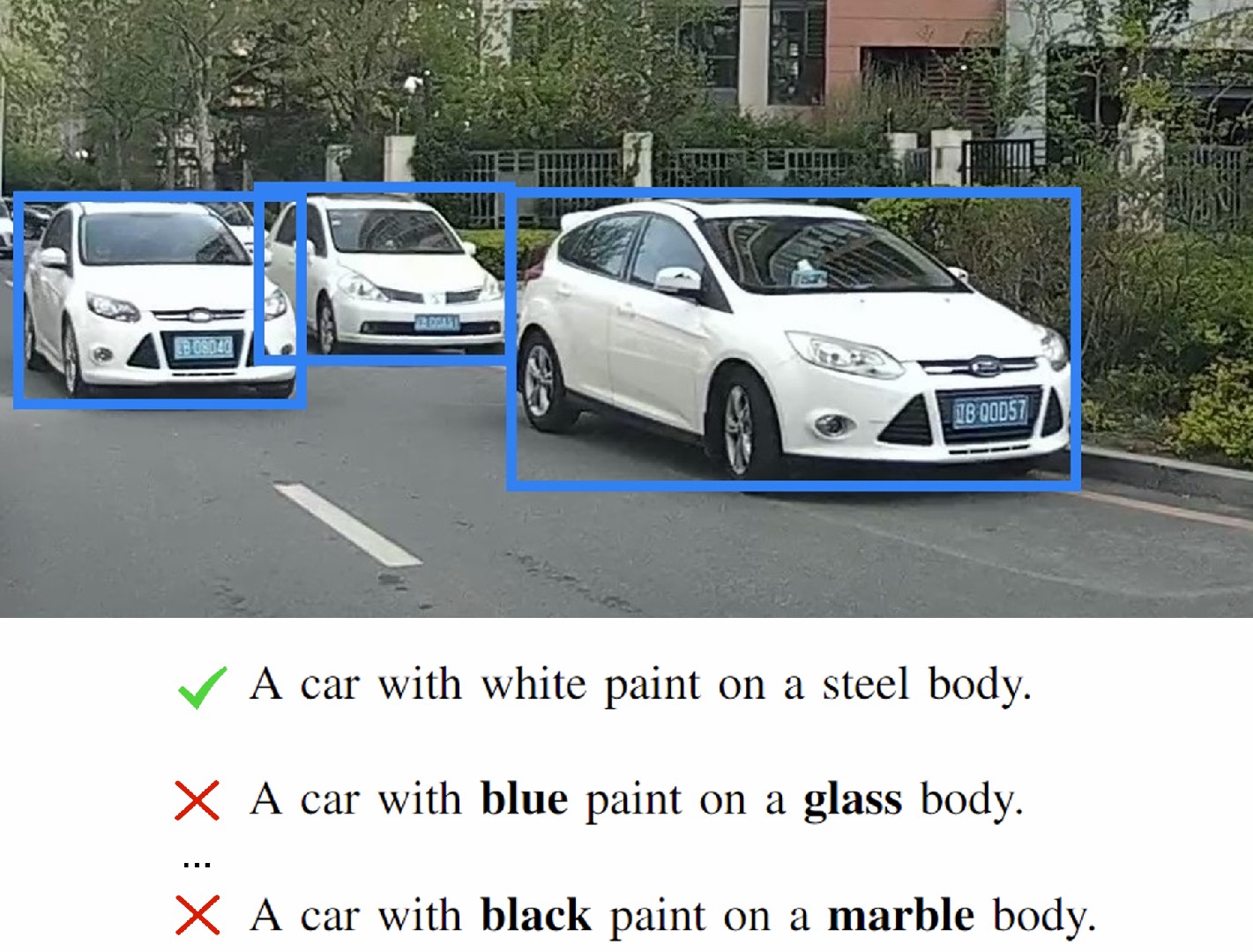}
         \caption{FG-OVD \cite{bianchi2024devil}.}
         \label{fig:fg-ovd}
     \end{subfigure}
     \begin{subfigure}[b]{0.23\textwidth}
         \centering
         \includegraphics[width=\textwidth]{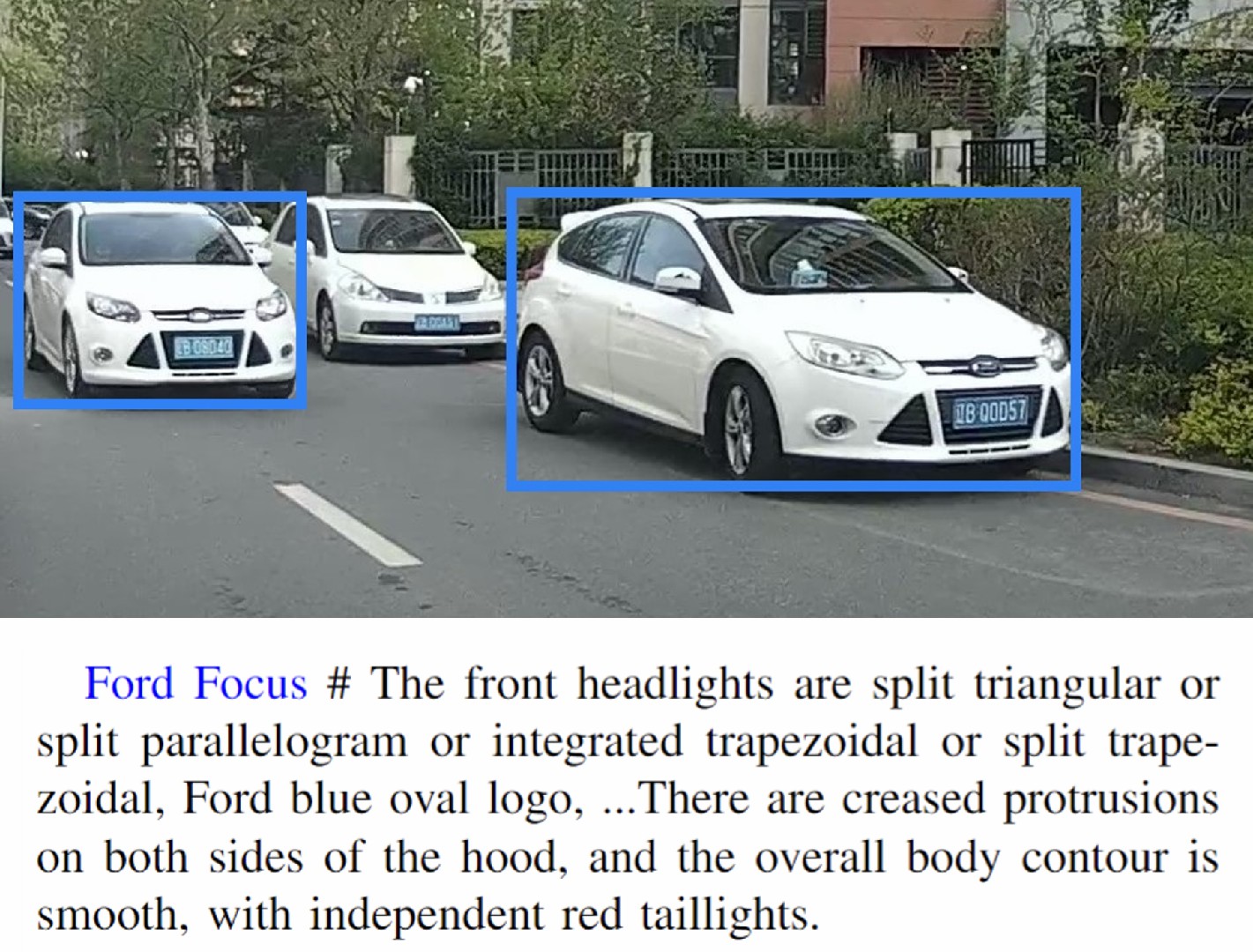}
         \caption{3F-OVD (ours).}
         \label{fig:3f-ovd}
     \end{subfigure}
    \caption{\textbf{Ground truth annotations for different evaluation tasks.} \textcolor{red}{Base classes} are highlighted in red while \textcolor{blue}{novel classes} are in blue. The FG-OD is the abbreviation of supervised Fine-Grained Object Detection. From the left to the right, the fine-grained classes of the cars are: ``Ford Focus'', ``Nissan Eida'', ``Ford Focus''.}
    \label{fig:task settings}
\end{figure}

OV-VG \cite{wang2024ov}
 increases the complexity of the task by incorporating attributes and relative locations of objects into the captions, as depicted in Fig.~\ref{fig:ov-vg}. This approach requires algorithms to comprehend detailed and extensive captions to predict correctly, even if novel classes are present in the pre-training data. However, because the task deviates from object detection to visual grounding, only one bounding box is predicted for each caption.
 On the other hand, FG-OVD \cite{bianchi2024devil} focuses solely on the attributes of objects in the captions, allowing multiple bounding boxes generated from a single caption. Nevertheless, this evaluation protocol asks for several false captions per each true caption to grant false positive predictions. Another limitation is that attributes (e.g., colors, material) are not always necessary for object detection; for instance, the three cars belonging to two different classes are all white in Fig.~\ref{fig:fg-ovd}. In summary, both evaluation protocols need image-level annotation for captions and could not reflect accurate capability of open-vocabulary detectors. A formal analysis will be provided in Sec.~\ref{sec:Evaluation Protocols}.



In this paper, we introduce an intuitive yet challenging evaluation task for open-vocabulary object detection. In our task, the novel classes are fine-grained classes instead of general classes, which can mitigate the risk of data leakage. In addition, a fine-grained class correlates with a caption, which demands for much less annotation. Apart from that, our task is more challenging. On one hand, because fine-grained objects exhibit small inter-class variations and large intra-class variations, open-vocabulary detectors must be capable of attending to subtle visual details in order to perform well in this task. For example, the three cars in Fig.~\ref{fig:3f-ovd} are very similar with nuances on the front. On the other hand, our captions are longer and more complex for detectors to digest.

Theoretically, as a natural extension of supervised fine-grained object detection, existing datasets can be adapted to our task by creating captions for each class. However, we found that the few existing fine-grained detection datasets are either too small or have certain limitations, as detailed in Sec.~\ref{sec:Characteristics}. To address these issues, we collected and annotated a novel fine-grained detection dataset, termed NEU-171K, for both classical object detection and open-vocabulary detection.

We also found, due to the increased complexity of our captions and the inclusion of more objects, existing open-vocabulary detectors often produce a higher number of false negatives, particularly in detecting local components of objects. To mitigate this issue, we propose a simple post-processing trick to enhance the performance of these detectors.

We summarize the contributions of our work as follows:
\begin{itemize}

\item We introduce a challenging evaluation task that fairly assesses open-vocabulary object detectors while mitigating data leakage risks and minimizing annotation overhead. The evaluation protocol is detailed in Sec.~\ref{sec:Evaluation Protocols}.

\item 
We release NEU-171K, the first large-scale fine-grained object detection dataset designed for both supervised and open-vocabulary settings. It contains 143,205 images, 659,286 bounding boxes, and 719 fine-grained classes, spanning two distinct domains. Dataset details are provided in Sec.~\ref{tab:DATASETS}. We benchmark state-of-the-art object detectors on both supervised and open-vocabulary detection, as described in Sec.~\ref{sec:EXPERIMENTS}.

\item We propose a simple yet effective post-processing trick to enhance the performance of existing open-vocabulary detectors by reducing false positives. The methodology is discussed in Sec.~\ref{sec:Post-processing}.

\end{itemize}

\section{RELATED WORK}
\subsection{Zero-Shot and Open-Vocabulary Detection}

Zero-shot object detectors \cite{antonelli2022few}, which can identify unknown classes without the need for expensive annotation, are less commonly used than their supervised counterparts due to their significantly lower performance. \cite{bansal2018zero} pointed out the missing background classes of region proposals is a challenge, because the background classes provide supervised object detection as the negative signals. Since every object is unseen in the training set, it is essentially a potential positive label. As a consequence, the zero-shot object detectors usually employ two stages approach, which consist of a generalized object proposal and object classifiers.

Since the zero-shot object detection has to utilize auxiliary data, \cite{zareian2021open} proposed the Open-Vocabulary object Detection (OVD), where image-caption pairs are also provided. The setting is more realistic because it imitates the natural supervision of human abilities and the image-caption pairs are much easier to access. The methods of OVD can be divided into five groups: knowledge distillation, region text pre-training, training with more balanced data, prompting modeling, and region text alignment.


\subsection{Visual Grounding and Phrase Localization}
Visual grounding, also known as referring expression comprehension, is the task of linking specific natural language expressions to corresponding regions within an image \cite{liu2019learning}. This task is crucial for enabling models to understand and interpret visual content based on human language, requiring a detailed understanding of both visual and linguistic information. Traditional approaches to visual grounding often relied on holistic sentence embedding methods, where the entire sentence is encoded as a single vector and matched with image regions. While simple, these methods often struggled with complex sentences and lost important compositional details.

Phrase localization is the task of identifying and localizing specific objects or regions within an image that correspond to a given textual phrase. Methods for phrase localization can be categorized into fully supervised approaches, which rely on paired phrase-region annotations (e.g., Canonical Correlation Analysis) \cite{plummer2017phrase}; weakly supervised methods, which use image-level annotations without explicit region mappings, often employing attention mechanisms or ranking models \cite{plummer2017phrase}; and unsupervised or non-paired methods, which utilize pre-trained detectors and semantic similarities to perform localization without paired training data \cite{plummer2017phrase}.

Ov-VG \cite{wang2024ov} employs visual grounding to evaluate the open-vocabulary object detectors. However, since only one bounding box is generated from a caption, the task differs from the vanilla object detection task, which could product multiple bounding boxes for a class.

\subsection{Attribute Detection and FG-OVD}
Attribute detection involves recognizing and classifying objects based on their specific visual properties, such as color, texture, shape, or material. Recent works like \cite{pham2021learning} introduced large-scale attribute detection datasets 
 annotated with both positive and negative attributes to facilitate learning and evaluation in realistic scenarios. \cite{bravo2023open} further extended attribute detection to an open-vocabulary setting, focusing on predicting attributes for object instances without explicitly annotated training examples, relying on zero-shot learning through vision-language models. 
 
 Fine-Grained Open-Vocabulary Detection (FG-OVD) \cite{bianchi2024devil} shares conceptual similarities with attribute detection, as both tasks emphasize detailed visual understanding. However, FG-OVD specifically aims to differentiate visually similar but categorically distinct objects, using fine-grained attributes as distinguishing factors. Thus, FG-OVD inherently relies on attribute detection to perform accurate visual discrimination, especially in cases where object identity is closely tied to attributes.




\subsection{Fine-Grained Datasets}
Because of the pressing interest and significant importance of fine-grained image analysis, various datasets for image classification and image retrieval have been released since 2008 \cite{nilsback2008automated} covering various domains, including aircraft, food, animals, vegetables, multi-modal, scenes and even sketches. Details of the popular benchmark datasets are summarized in \cite{wei2021fine}.

The captions for our task correlate can be generated from fine-grained classes, however, limited fine-grained detection datasets are public available, and existing ones have a variety of shortcomings. We will reveal more details in Sec. \ref{sec:Characteristics}. It is worth noting that although some datasets \cite{bai2020products, khosla2011novel} provide bounding boxes, they categorize themselves as image classification dataset because each image only contains one object. Therefore, we created NEU-171K dataset to benefit the research for both supervised fine-grained object detection and open-vocabulary object detection.


\section{METHODOLOGY}

\subsection{Evaluation Protocols} \label{sec:Evaluation Protocols}
Let us denote a dataset of \( N \) images as \(\{I_i\}_{i=1}^N\), where each image \( I_i \) is associated with annotations \( \{(b_{i,j}, c_{i,j})\}_{j=1}^{m_i} \). Here, \(m_i\) is the number of annotated bounding boxes for the \(i\)-th image. The notation \(b_{i,j} \in \mathbb{R}^4\) represents the \(j\)-th bounding box, and \(\ell_{i,j} \in \{1, \ldots, K\}\) is the label of that bounding box chosen from a closed set of \(K\) classes.

A traditional supervised object detector \(\psi\) attempts to learn from these images and annotations so that, at inference time, given a new image \(I\), it produces
\(\psi(I) = \bigl\{(b_j^{\text{pred}}, s_j^{\text{pred}}, c_j^{\text{pred}})\bigr\}_{j=1}^{\widehat{m}},\)
where:
\( \widehat{m} \) is the number of predicted boxes,
\( b_j^{\text{pred}} \in \mathbb{R}^4 \) is the predicted bounding box,
\( s_j^{\text{pred}} \in \mathbb{R}_{\ge 0} \) is the confidence score of that box,
\( c_j^{\text{pred}} \in \{1,\ldots,K\}\) is the predicted class label from the same closed set used in training.

In the open-vocabulary setting, we distinguish between base classes \(\mathcal{C}_{\text{base}}\) (seen during training) and novel classes \(\mathcal{C}_{\text{novel}}\) (unseen during training). The fundamental idea is that the training data includes bounding boxes only for \(\mathcal{C}_{\text{base}}\), but during inference, the detector should handle classes from \(\mathcal{C}_{\text{base}} \cup \mathcal{C}_{\text{novel}}\). 
Mathematically, let \(\mathcal{C} = \mathcal{C}_{\text{base}} \cup \mathcal{C}_{\text{novel}}\) be the total set of classes, and \(\mathcal{V}\) be a large vocabulary of words (or text embeddings) describing all these classes (and potentially more). During inference, an open-vocabulary detector \(\psi_{\text{OV}}\) is provided with:
1. An image \(I\).
2. Class name $c^{\text{pred}}$.
3. (Optionally) a caption or visual description \(\{\mathbf{t}_c : c \in \mathcal{C}\} \subset \mathcal{V}\),

Then the detector produces predicted bounding boxes and scores:
\(
    \psi_{\text{OV}}\bigl(I, \{\mathbf{t}_c\}_{c \in \mathcal{C}}, c^{\text{pred}})
    \;=\;
    \bigl\{ (b_j^{\text{pred}}, s_j^{\text{pred}}, \ell_j^{\text{pred}}) \bigr\}_{j=1}^{\widehat{m}},
\)
where \(\ell_j^{\text{pred}} \subset \mathcal{V}\ \) is a word or token from the caption. The captions \(\mathbf{t}_c\) serve as part of the classification mechanism to allow recognition of classes not explicitly annotated in the training phase.

Traditional Open-Vocabulary Visual Grounding (OV-VG) \cite{xiao2024towards,wang2024ov} focuses on localizing a specific phrase or sentence in an image. Given an image \(I_i\) and a caption \(p_i\in \mathcal{V}\) describing an object, a grounding algorithm must output exactly one bounding box that aligns with that caption. Let us denote the caption by \(\mathbf{d}\) (a feature vector from the text encoder) and the bounding box by \(\hat{b}\in \mathbb{R}^4\). In many OV-VG tasks, the caption can vary per image and often includes additional information like attributes, relationships, etc. Mathematically, we can write:
\(
   \hat{b} = \arg\max_{b}\; \mathrm{Score}(\phi(I_i, b), \mathbf{d}),
\)
where \(\phi(I, b)\) extracts the visual representation (e.g., cropped features) for box \(b\) from \(I\), and \(\mathrm{Score}(\cdot, \cdot)\) measures alignment between visual features and textual features. In OV-VG, each caption or phrase can be unique per image, so captions for classes can differ from image to image. While in our task, as described below, the same class description (caption) is re-used across images.

Fine-grained open-vocabulary detection (FG-OVD) \cite{bianchi2024devil} emphasizes that each object instance is accompanied by a set of one positive caption describing that instance (with fine-grained attributes) and several negative captions that differ slightly from the positive one. Concretely, for an image \(I_i\), suppose there are \(m\) objects \( \{o_j\}_{j=1}^m\). In FG-OVD, each object \(o_j\) belongs to a base or novel class \(c_j \in \mathcal{C}\). It constructs a group of captions \(\mathcal{P}_j = \{\mathbf{t}_j^{+}, \mathbf{t}_{j,1}^{-}, \ldots, \mathbf{t}_{j,K}^{-}\}\) for each object \(o_j\), where \(\mathbf{t}_j^{+}\) is the positive (correct) caption for object \(o_j\),
and \(\{\mathbf{t}_{j,k}^{-}\}\) are negative captions.

The FG-OVD system must produce
\(
   \bigl\{(b_j^{\text{pred}}, s_j^{\text{pred}}, c_j^{\text{pred}})\bigr\}_{j=1}^{m}
\)
subject to each bounding box \(b_j^{\text{pred}}\) matching the correct ground-truth bounding box for the object. Meanwhile, \(c_j^{\text{pred}}\) must be inferred from the group of captions \(\mathcal{P}_j\). As discussed in \cite{bianchi2024devil}, the challenge is that each image instance has its own group of captions, making the search space dynamic per object.

Finally, in our open-vocabulary detection setting (3F-OVD), each class \(c\in \mathcal{C}\) has a single caption \(\mathbf{t}_{c}\) re-used for all images. At inference, the model is presented with an image \(I_i\) and a textual description set \(\{\mathbf{t}_c\}_{c \in \mathcal{C}}\), which is constant across images, and must output
\(
    \psi_{\text{ours}}(I) \;=\; \bigl\{(b_j^{\text{pred}}, s_j^{\text{pred}}, \ell_j^{\text{pred}})\bigr\}_{j=1}^{\widehat{m}},
\)
\(\ell_j^{\text{pred}} \subset \mathcal{V}\ \) is a word or token from the caption. Importantly, the same caption or class description \(\mathbf{t}_c\) is shared by all images for the class \(c\), which differs from OV-VG (where each image-phrase pair can be unique). In our approach, we also store \(\widehat{m}\) confidence scores \(\{s_j^{\text{pred}}\}\) that measure alignment between the box-level features and class text embeddings. The ability to handle novel classes arises from learning text–image alignment during training on \(\mathcal{C}_{\text{base}}\) and leveraging the class description \(\mathbf{t}_c\) for any class \(c\), even if unobserved previously.

Although the time complexity of our evaluation protocol is $(\alpha + \beta) O(NK)$, it can be reduced to $\alpha O(N) + \alpha O(K) + \beta O(NK)$, where $\alpha$ is the time coefficient for encoders and $\beta$ is the time coefficient for alignment, if we pre-compute text embeddings for captions.

\subsection{Post-processing}
\label{sec:Post-processing}

\begin{figure}[!htbp] 
    \centering
    \begin{subfigure}[b]{0.23\textwidth} 
        \centering
        \includegraphics[width=\textwidth]{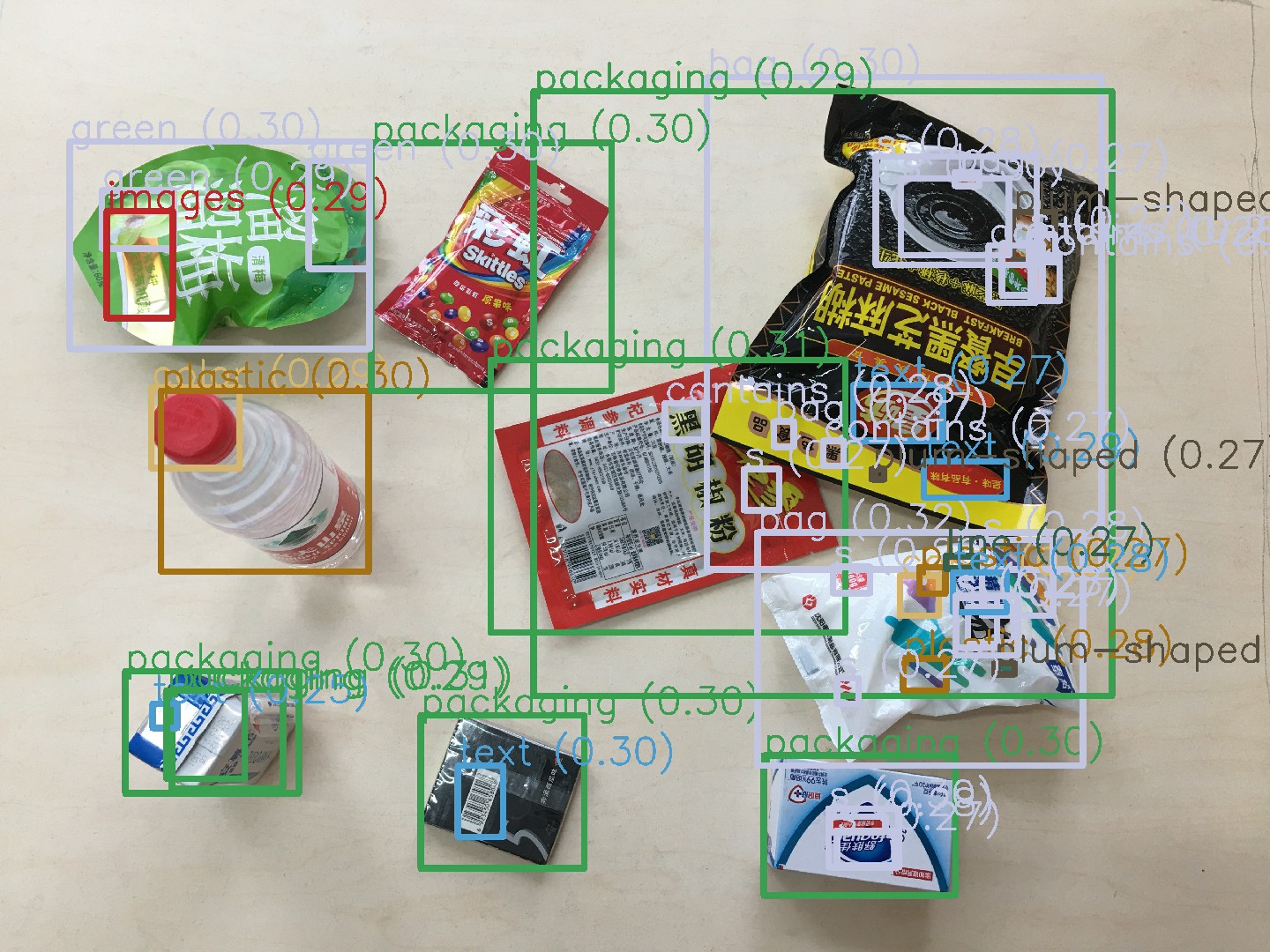}
        \caption{Before our post-processing trick.}
        \label{fig:raw_rb_before}
    \end{subfigure}
    \vspace{1em} 
    \begin{subfigure}[b]{0.23\textwidth}
        \centering
        \includegraphics[width=\textwidth]{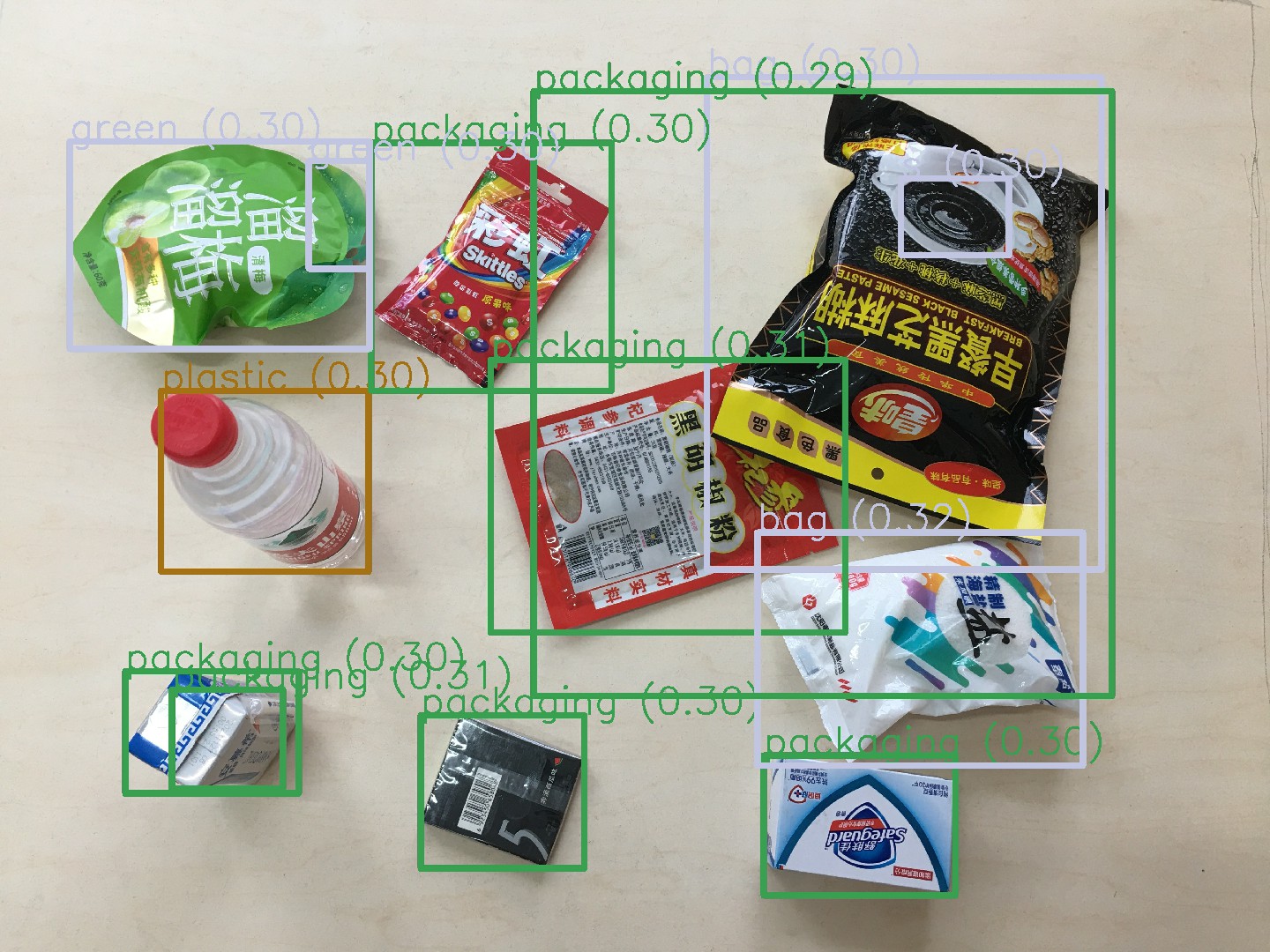}
        \caption{After our post-processing trick.}
        \label{fig:raw_rb_after}
    \end{subfigure}
    \caption{\textbf{Bounding boxes predicted by ViLD on NEU-171K-RP before (Fig.~\ref{fig:raw_rb_before}) and after (Fig.~\ref{fig:raw_rb_after}) applying our post-processing method.} The evaluated class is ``Dried Fruit Liuliumei Green'' with the caption: ``The packaging of Liuliumei Green is a plum-shaped plastic bag predominantly green ... and the lower-right corner contains a line of small text.''. Each bounding box is labeled with the corresponding word and confidence score.}
    \label{Fig:pp_visualize}
\end{figure}

We observed three key phenomena in the predictions of open-vocabulary detectors:
\begin{enumerate*}[font={\bfseries}]
\item Class names are often not included in their vocabulary, as seen in the caption ``Dried Fruit Liuliumei Green'' in the Fig.~\ref{Fig:pp_visualize}. The tokenizers of detectors could not find a text embedding for the class name or give a special token [UNK].
\item Captions, being visual descriptions, often generate multiple false-positive bounding boxes for individual words. For example, in Fig.~\ref{fig:raw_rb_before}, the word ``text'' from the caption generates small bounding boxes inside products that contain text.
\item Current open-vocabulary object detectors cannot predict at the caption level, but only at the word or token level.
\end{enumerate*}

Based on these observations, we propose a post-processing trick that aims to eliminate false-positive bounding boxes. This trick is similar to NMS but instead calculates the proportion of overlapping areas between candidate bounding boxes rather than using IoUs. Additionally, we eliminate bounding boxes that are excessively large or small.

\section{DATASET} \label{tab:DATASETS}

\subsection{Construction}
Our dataset consists of two domains: cars (NEU-171K-C) and retail products (NEU-171K-RP). The vehicle images were collected from real-world road traffic, while the retail product images were captured in a controlled lab setting to simulate warehouse conditions.

For NEU-171K-C, videos were collected using advanced handheld cameras and dashcams, yielding 89,363 images at a resolution of $1080 \times 1920$ pixels. To prevent data leakage, we ensured that any two frames captured within 5 seconds of each other are always assigned to the same set. If two frames are more than 5 seconds apart, they may be placed in the same set or different sets without restriction.

In contrast, NEU-171K-RP is visually simpler for detectors, as retail products exhibit clearer visual distinctions and their backgrounds remain static. This setup enables us to study the effect of fine-grained visual information. This dataset includes 53,842 images at a resolution of $3024 \times 4032$ pixels. 

\begin{figure}[!h]
    \centering
    \begin{subfigure}[b]{0.48\textwidth}
         \centering
         \includegraphics[width=\textwidth]{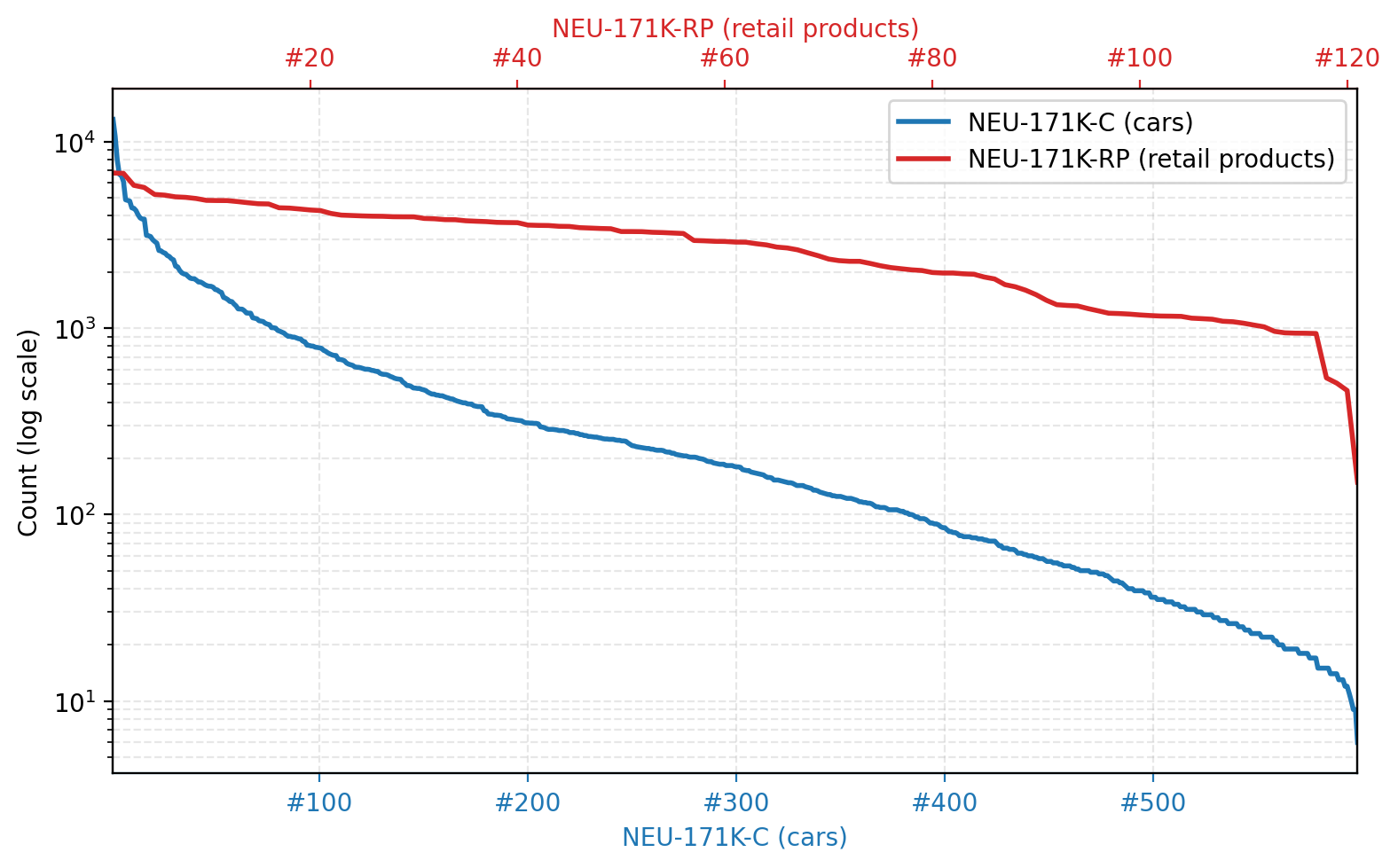}
     \end{subfigure}
    \caption{\textbf{Class distribution of images in NEU-171K-RP (red) and NEU-171K-C (blue).} The X-axis represents class IDs, sorted in descending order by frequency, while the Y-axis indicates the number of images per class.}
    \label{fig:distribution}
\end{figure}

The distributions of both datasets are illustrated in Fig.~\ref{fig:distribution}. From the figure, we can see that NEU-171K-C follows a more pronounced long-tailed distribution compared to NEU-171K-RP, aligning with their respective environmental conditions.

\subsection{Annotation}
We used LabelMe \cite{torralba2010labelme} for the annotation. Our dataset includes 598 fine-grained classes for NEU-171K-C and 121 for NEU-171K-RP. To ensure precise annotations, we assumed that each object was composed of multiple parts, with each part characterized by at least one attribute. If more than two-thirds of an object's estimated area was occluded by another object, we did not assign a bounding box to it, as determining its class under such conditions was impractical. For each class, we first composed a caption in Chinese and then translated it into English using ChatGPT. 


\subsection{Characteristics}
\label{sec:Characteristics}
\begin{table}[htp]
\begin{center}
\begin{threeparttable}

  \begin{tabular}[h]{ | c  c  c  c  c  c |}
    \hline
    Dataset & Img & Cls & Sub & Bb & Task \\ \hline
    FGVD\cite{khoba2022fine} & 5,502 & / & 217 & 24,450 & D \\
    Military Aircraft\cite{nakamura2020mad} & 15,948 & / & 60 & 27,564 &D \\
    AC\cite{Steininger2021AircraftContextDataset} & 42,000 & / & 4,000 & 51,000 & D \\
    RPC\cite{wei2022rpc} & 83,739\tnote{*} & / & 200 & 156,774 & D \\
    iNaturalist2017\cite{van2018inaturalist} & 675,170 & 13 & 5,089 & 561,767 & D \\ \hline
    LVIS\cite{gupta2019lvis} & 119,979 & 1,203 & / & 1,514,848 & O \\
    OV-COCO\cite{zareian2021open} & 123,287 & 80 & / & 837,500 &O \\ \hline
    NEU-171K-C & 89,363 & 88 & 598 & 316,094 & O/D \\
    NEU-171K-RP & 53,842 & 17 & 121 & 343,192 & O/D \\
    NEU-171K & 143,205 & 105 & 719 & 659,286 & O/D \\ 
    \hline
  \end{tabular}
  \begin{tablenotes}
        \footnotesize
        \item[*] The training set of the RPC dataset comprises 53,739 single-object samples, where each image contains one object.       In contrast, the collective validation set and testing set consist of 30,000 multi-object detection samples.  
      \end{tablenotes}

  \end{threeparttable}

  \caption{Comparison of representative fine-grained recognition datasets. \#Img (images), \#Cls (classes), \#Sub (sub-classes), and \#Bb (bounding boxes) indicate dataset statistics. D, and O denote the tasks of object detection, and open-vocabulary detection, respectively.}
  \label{tab: data-chara}

\end{center}
\end{table}

A dataset is considered suitable for fine-grained object detection only if it contains multiple bounding boxes per image and the majority of its classes exhibit fine-grained distinctions. Therefore, COCO \cite{lin2014microsoft} does not qualify as a fine-grained object detection dataset, as the majority of its classes lack this characteristic.

We compare our dataset with other similar datasets in Tab.~\ref{tab: data-chara}. The FGVD \cite{khoba2022fine}and Military Aircraft \cite{nakamura2020mad} datasets contain a relatively small number of images. While the AC dataset \cite{Steininger2021AircraftContextDataset} contains 42,000 images, each sub-class has relatively few images, given that there are 4,000 sub-classes. iNaturalist2017 \cite{van2018inaturalist} has the largest number of images among these datasets, but most images contain only one sub-class. RPC \cite{wei2022rpc} contains the most images among these datasets, but in most images, multiple objects are present, all belonging to the same sub-class.

OV-COCO \cite{zareian2021open} and LVIS \cite{gupta2019lvis} are the two most widely used open-vocabulary object detection datasets. Our dataset can also be used for this task, offering more images and fine-grained classes.

\section{EXPERIMENTS} \label{sec:EXPERIMENTS}

\subsection{Classical Object Detection}
We benchmarked state-of-the-art supervised object detectors on our datasets and compared their performance against other datasets. Each fine-grained class was included in the training, validation, and testing sets. We used standard Non-Maximum Suppression (NMS) for post-processing and COCO mAP as the evaluation metric. The number of images allocated for training, validation, and testing was 64658:12903:11802 in NEU-171K-C and 34417:8639:10786 in NEU-171K-RP.

\textbf{Methods.} We benchmarked seven popular object detectors: Co-detr \cite{zong2023detrs}, FCOS \cite{tian2020fcos}, CenterNet \cite{zhou2019objects}, PAA \cite{kim2020probabilistic}, Faster R-CNN \cite{ren2016faster}, EfficientNet \cite{tan2021efficientnetv2} and GFL \cite{li2020generalized}. We followed the default training configurations as specified in the original publications of each model. Unless otherwise specified, we used ResNet-50 as the backbone for these object detectors, and the hyperparameters remained consistent with the settings in the RetinaNet framework.

\textbf{Analysis.} 
As shown in Tab.~\ref{tab: classic-object-detection-ap}, detectors performed better on RPC \cite{wei2022rpc} and NEU-171K-RP datasets than on COCO dataset, as both contain retail products only. Since NEU-171K-RP was created in a controlled environment with clearly distinguishable objects, all detectors demonstrated the highest accuracy among the tested datasets. In contrast, the detectors performed worse on the FGVD and NEU-171K-C datasets compared to COCO, as both focus on vehicle detection, more challenging datasets due to high intra-class variation and complex backgrounds.

\begin{table}[htp]
\begin{center}
   \scalebox{0.83}{
  \begin{tabular}[h]{ | c | c c | c | c c | }
    \hline
    & \multicolumn{2}{|c|}{Car} & Common Obj & \multicolumn{2}{|c|}{Retail Products} \\ \hline
    Detector & FGVD & NEU-171K-C & COCO & RPC & NEU-171K-RP \\ \hline
    PAA & 0.241 & 0.335 & 0.411 & 0.878 & 0.954 \\
    FasterRCNN & 0.252 & 0.247 & 0.348 & 0.680 & 0.948 \\
    EfficientNet & 0.128 & 0.240 & 0.361 & 0.849 & 0.958 \\
    FCOS & 0.084 & 0.314 & 0.389 & 0.864 & 0.951 \\
    CenterNet & 0.096 & 0.352 & 0.421 & 0.871 & 0.951 \\
    GFL & 0.056 & 0.328 & 0.387 & 0.878 & 0.962 \\
    Co-detr & 0.274 & 0.387 & 0.495 & 0.878 & 0.955 \\

    \hline
  \end{tabular}
  }
\begin{tablenotes}
        \footnotesize
        \item* The RPC dataset used here consists of only the validation and test sets (i.e., the multi-objective part).  
\end{tablenotes}
  \caption{mAP of state-of-the-art supervised object detectors across multiple datasets.}
  \label{tab: classic-object-detection-ap}
\end{center}
\end{table}

\subsection{Open-Vocabulary Object Detection} \label{exp:Open-Vocabulary Object Detection}
In this section, we compare our evaluation protocol with the other two evaluation protocols: FG-OVD and OV-VG. Note that the same dataset cannot be used for different evaluation protocols because our captions are class-specific rather than image-specific.


\textbf{Methods.} Our benchmark includes three state-of-the-art detectors: ViLD \cite{gu2021open}, Detic \cite{zhou2022detecting}, and GroundingDino (GDino) \cite{liu2023grounding}. Specifically, GroundingDino uses the Swin Transformer as its visual backbone, the BERT text encoder for textual representations, and was pre-trained on the O365, Gold G, and Cap4M datasets. We do not provide token spans for GroundingDino. In other words, each word is considered an individual class. Detic employs the Swin Transformer as its visual backbone. Its text encoder uses CLIP, with ViT-B/32 as its pretrained weight configuration, and it was pre-trained on the LVIS+COCO and ImageNet-21K datasets. ViLD employs ResNet-50 as its visual backbone, relies on the CLIP text encoder for textual representations, and was pre-trained on the OV-COCO dataset.

For FG-OVD, we used ten easy dynamic vocabularies (N=10) and applied its color attribute replacement to evaluate the open-vocabulary detector's ability to detect changes in target attributes. In addition, we followed their default model configuration and pre-trained weights.

Since the complete annotation files for OV-VG were not publicly available, we reconstructed the data using the officially released Open-Vocabulary Phrase Localization (OV-PL) subset, which was obtained via stratified sampling from the OV-VG dataset. The data reconstruction process consisted of three key steps:
\begin{enumerate*}[font={\bfseries}]
\item Image Description Adjustment: We modified the original global descriptions of images, converting them into referential descriptions focused on specific objects.
\item Class Filtering: Rare-class bounding boxes were removed, and only base-class annotations corresponding to captions were retained.
\item Description-Annotation Alignment: We implemented a semantic matching mechanism to ensure that the retained bounding boxes corresponded precisely to the reconstructed referential descriptions.
\end{enumerate*}
If the first matched description for an image exhibits multi-object referential ambiguity, the entire sample is discarded to maintain annotation reliability.

\textbf{Analysis.} As shown in Tab.~\ref{tab: data-chara3}, all open-vocabulary detectors perform significantly worse on 3F-OVD (NEU-171K-C and NEU-171K-RP) than FG-OVD and OV-VG, indicating it as the most challenging evaluation protocol among those tested. Similar to the Tab.~\ref{tab: classic-object-detection-ap}, performance on NEU-171K-C was generally better than on NEU-171K-RP, as cars exhibit greater visual similarity and more complex backgrounds than retail products.
However, the performances on NEU-171K-RP are not always the best, indicating that both visual detail and language understanding are critical for effective open-vocabulary detection.

\begin{table}[htp]
\begin{center}
\scalebox{0.94}{
  \begin{tabular}[h]{ | c | c | c | c | c | c |}
    \hline
    Method & FG-OVD & OV-VG & NEU-171K-C & NEU-171K-RP \\ \hline
    Detic & \num{6.6e-02} & \num{2.1e-01} & \num{6.3e-04} & \num{2.0e-02} \\
    GDino & \num{2.2e-01} & \num{8.0e-01} & \num{1.2e-03} & \num{7.4e-04} \\
    ViLD & \num{2.6e-01} & \num{1.5e-01} & \num{3.3e-04} & \num{7.5e-03} \\ 
    \hline
  \end{tabular}
  }
  \caption{mAP of open-vocabulary object detectors under three evaluation protocols: FG-OVD, OV-VG, and 3F-OVD. Both NEU-171K-C and NEU-171K-RP are evaluated using the 3F-OVD protocol.}
  \label{tab: data-chara3}
\end{center}
\end{table}

\subsection{Post-process}
\begin{figure}[h]
     \centering
     \begin{subfigure}[b]{0.23\textwidth}
         \centering
         \includegraphics[width=\textwidth]{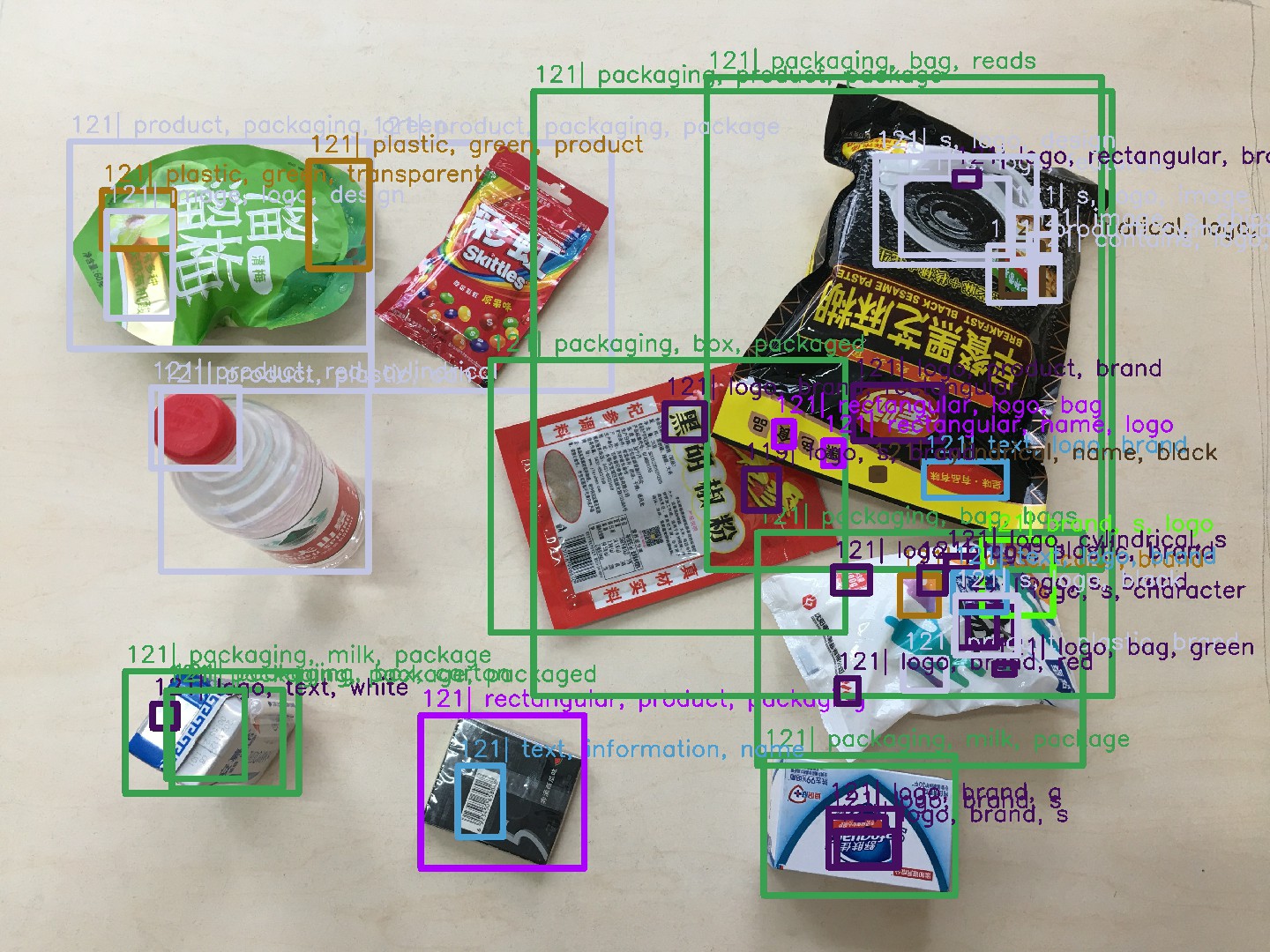}
         \caption{Without our post-processing trick on NEU-171K-RP.}
         \label{GDino-rp}
     \end{subfigure}
     \begin{subfigure}[b]{0.23\textwidth}
         \centering
             \includegraphics[width=\textwidth]{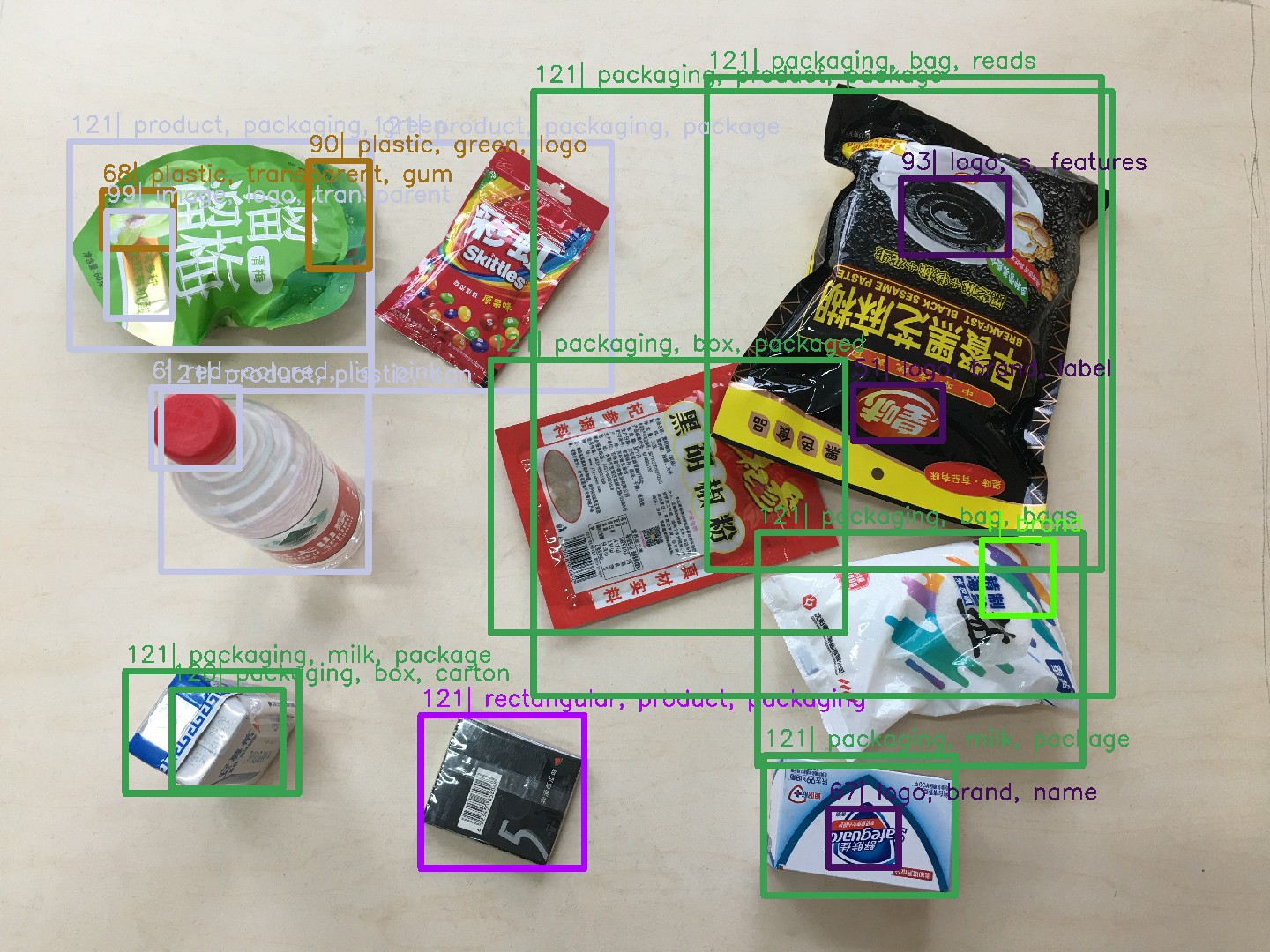}
         \caption{With our post-processing trick on NEU-171K-RP.}
         \label{GDino-rp-trick}
     \end{subfigure}
     \begin{subfigure}[b]{0.23\textwidth}
         \centering
         \includegraphics[width=\textwidth]{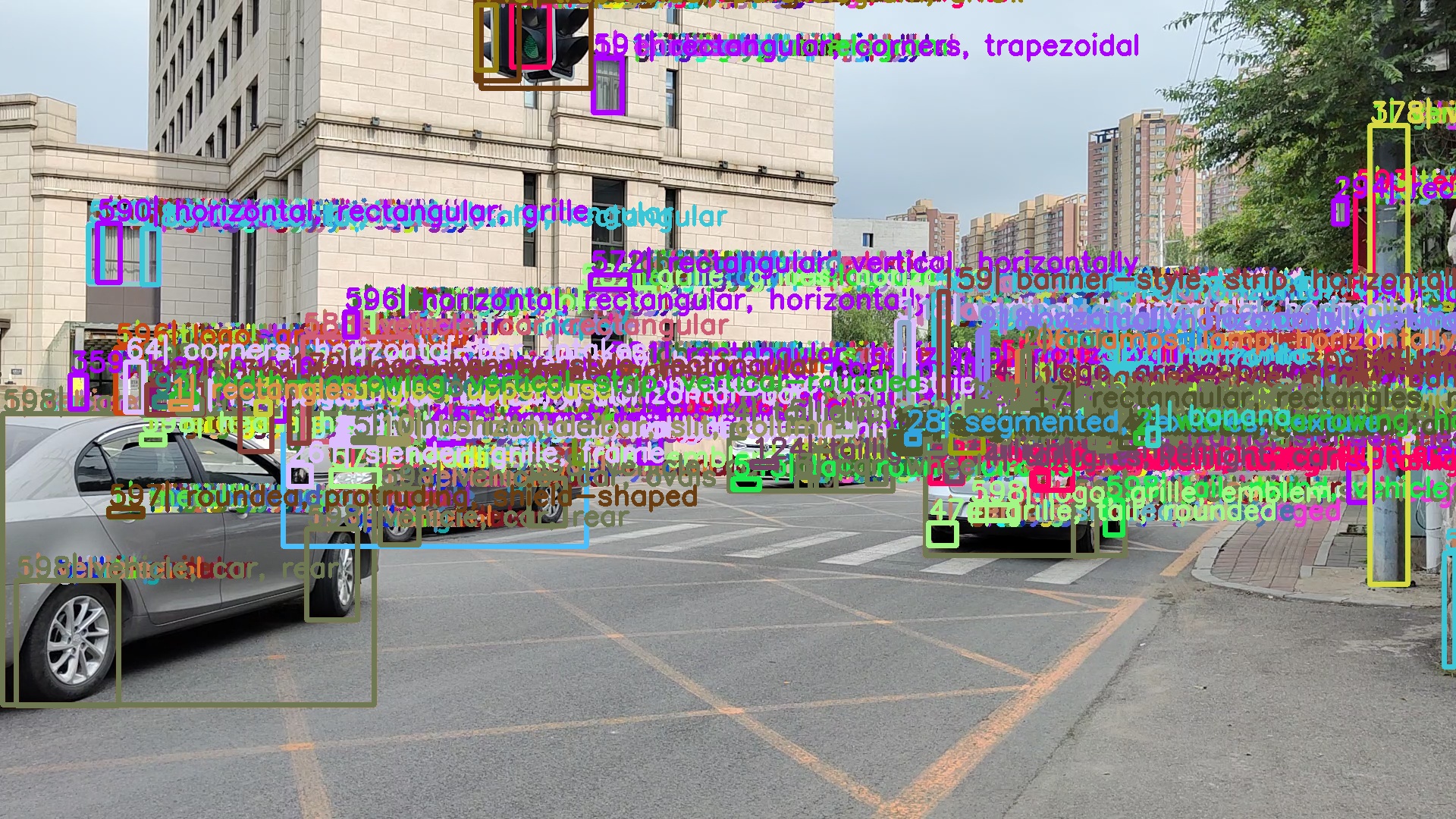}
         \caption{Without our post-processing trick on NEU-171K-C.}
         \label{GDino-c}
     \end{subfigure}
     \begin{subfigure}[b]{0.23\textwidth}
         \centering
             \includegraphics[width=\textwidth]{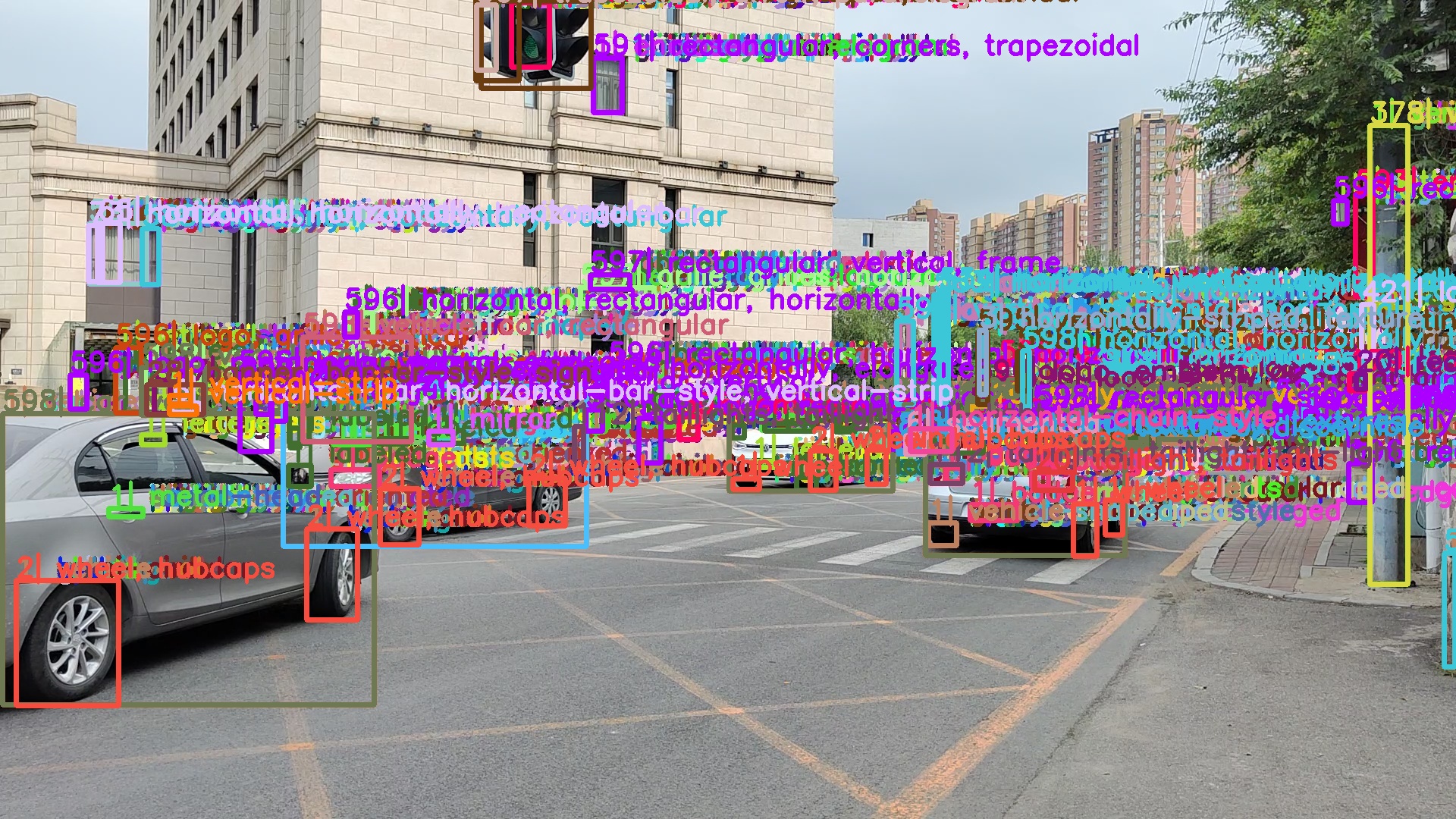}
         \caption{With our post-processing trick on NEU-171K-C.}
         \label{GDino-c-trick}
     \end{subfigure}
    \caption{\textbf{Predicted bounding boxes from ViLD before and after applying our post-processing method across all classes.} Each bounding box is labeled with the number of occurrences and the three most frequently predicted words.}
    \label{Fig:Ablation study}
\end{figure}
In this section, we conduct experiments to validate the performance of our post-processing trick. The open-vocabulary detectors used in this section adopt the same settings from Sec.~\ref{exp:Open-Vocabulary Object Detection}.

\textbf{Methods.}
Our post-processing method is applied to each caption individually, and the predictions from all captions within an image are aggregated. 
If a bounding box overlaps more than 80\% of its total area with another box that has a higher confidence score, it is removed. Additionally, we remove bounding boxes with areas smaller than $200 \times 200$ pixels or larger than $2250 \times 2000$ pixels for NEU-171K-RP. For NEU-171K-C, we eliminate boxes smaller than $14 \times 14$ pixels or larger than $960 \times 960$ pixels. All hyper-parameters in our post-processing method were determined empirically.



\textbf{Analysis.} Fig.~\ref{Fig:Ablation study} shows the bounding boxes of with and without our post-processing trick on both sets of our dataset. Since different captions can generate identical bounding boxes for an image, we annotate each bounding box with the number of times it was predicted. For example, if a bounding box is labeled with ``121'', it indicates that all 121 captions generated the same bounding box.

Fig.~\ref{Fig:Ablation study} shows that many small bounding boxes within object bounding boxes are removed, significantly reducing false positive predictions. For example, Fig.~\ref{GDino-rp} and Fig.~\ref{GDino-rp-trick} show that bounding boxes related to the words ``text'' and ``logo'' in NEU-171K-RP are removed. In NEU-171K-C, the bottom-left wheel, which initially had 598 bounding boxes, is reduced to just 2 after applying our post-processing method, as shown in Fig.~\ref{GDino-c} and Fig.~\ref{GDino-c-trick}.

As shown in Tab.~\ref{tab: data-chara4}, our post-processing method is robust, consistently improving performance for all open-vocabulary detectors on both NEU-171K-C and NEU-171K-RP. Intuitively, more sensitive detectors achieve greater performance improvements from our post-processing method, as it is more effective at eliminating false positive bounding boxes.


\begin{table}[!htp]
\begin{center}
    \begin{tabular}[h]{ | p{1.0cm} | p{0.5cm} | p{2.5cm} | p{2.5cm} |}
    \hline
    Method & trick & NEU-171K-C & NEU-171K-RP \\ \hline
    GDino  & w/o & \num{1.2e-03} & \num{7.4e-04} \\
    GDino  & w  & \num{1.3e-03}(+8.3\%) & \num{7.6e-04}(+2.6\%) \\  \hline
    Detic  & w/o & \num{6.3e-04} & \num{2.0e-02} \\ 
    Detic  & w  & \num{6.6e-04}(+4.7\%) & \num{2.2e-02}(+10.0\%) \\    \hline
    Vild   & w/o & \num{3.3e-04} & \num{7.5e-03} \\ 
    Vild   & w & \num{3.8e-04}(+15.2\%) & \num{10.6e-03}(+41.3\%) \\ 
    \hline
  \end{tabular}
  
  \caption{mAP with (w) and without (w/o) our post-processing method. The results show consistent performance improvements across all open-vocabulary detectors on NEU-171K-C and NEU-171K-RP.}

  \label{tab: data-chara4}
\end{center}
\end{table}

\section{Conclusion}

We introduced 3F-OVD, a novel and challenging task that extends supervised fine-grained object detection to an open-vocabulary setting. To support this, we developed NEU-171K, a large-scale dataset for both supervised and open-vocabulary detection. Our experiments demonstrate the challenges existing models face in this setting, and we propose an effective post-processing method to enhance detection accuracy. This work establishes a new benchmark for fine-grained open-vocabulary detection and opens the door for further research.






\section*{ACKNOWLEDGMENT}
We extend our gratitude to Professor Ye Lei and Professor Liu Guoqi's team for their support and assistance during the data annotation process. The authors also sincerely appreciate Mr. Diancheng Li for his help in writing, as well as Prof. Felipe Viana and Dr. Chih-Hui Ho for their instrumental assistance in the experiments.

\printbibliography

@article{wei2021fine,
  title={Fine-grained image analysis with deep learning: A survey},
  author={Wei, Xiu-Shen and Song, Yi-Zhe and Mac Aodha, Oisin and Wu, Jianxin and Peng, Yuxin and Tang, Jinhui and Yang, Jian and Belongie, Serge},
  journal={IEEE transactions on pattern analysis and machine intelligence},
  volume={44},
  number={12},
  pages={8927--8948},
  year={2021},
  publisher={IEEE}
}

@inproceedings{bianchi2024devil,
  title={The devil is in the fine-grained details: Evaluating open-vocabulary object detectors for fine-grained understanding},
  author={Bianchi, Lorenzo and Carrara, Fabio and Messina, Nicola and Gennaro, Claudio and Falchi, Fabrizio},
  booktitle={Proceedings of the IEEE/CVF Conference on Computer Vision and Pattern Recognition},
  pages={22520--22529},
  year={2024}
}

@inproceedings{nilsback2008automated,
  title={Automated flower classification over a large number of classes},
  author={Nilsback, Maria-Elena and Zisserman, Andrew},
  booktitle={2008 Sixth Indian conference on computer vision, graphics \& image processing},
  pages={722--729},
  year={2008},
  organization={IEEE}
}

@article{bai2020products,
  title={Products-10k: A large-scale product recognition dataset},
  author={Bai, Yalong and Chen, Yuxiang and Yu, Wei and Wang, Linfang and Zhang, Wei},
  journal={arXiv preprint arXiv:2008.10545},
  year={2020}
}

@inproceedings{van2018inaturalist,
  title={The inaturalist species classification and detection dataset},
  author={Van Horn, Grant and Mac Aodha, Oisin and Song, Yang and Cui, Yin and Sun, Chen and Shepard, Alex and Adam, Hartwig and Perona, Pietro and Belongie, Serge},
  booktitle={Proceedings of the IEEE conference on computer vision and pattern recognition},
  pages={8769--8778},
  year={2018}
}

@inproceedings{khosla2011novel,
  title={Novel dataset for fine-grained image categorization: Stanford dogs},
  author={Khosla, Aditya and Jayadevaprakash, Nityananda and Yao, Bangpeng and Li, Fei-Fei},
  booktitle={Proc. CVPR workshop on fine-grained visual categorization (FGVC)},
  volume={2},
  number={1},
  year={2011},
  organization={Citeseer}
}

@inproceedings{khoba2022fine,
  title={A Fine-Grained Vehicle Detection (FGVD) Dataset for Unconstrained Roads*},
  author={Khoba, Prafful Kumar and Parikh, Chirag and Jawahar, CV and Sarvadevabhatla, Ravi Kiran and Saluja, Rohit},
  booktitle={Proceedings of the Thirteenth Indian Conference on Computer Vision, Graphics and Image Processing},
  pages={1--9},
  year={2022}
}

@article{antonelli2022few,
  title={Few-shot object detection: A survey},
  author={Antonelli, Simone and Avola, Danilo and Cinque, Luigi and Crisostomi, Donato and Foresti, Gian Luca and Galasso, Fabio and Marini, Marco Raoul and Mecca, Alessio and Pannone, Daniele},
  journal={ACM Computing Surveys (CSUR)},
  volume={54},
  number={11s},
  pages={1--37},
  year={2022},
  publisher={ACM New York, NY}
}

@inproceedings{bansal2018zero,
  title={Zero-shot object detection},
  author={Bansal, Ankan and Sikka, Karan and Sharma, Gaurav and Chellappa, Rama and Divakaran, Ajay},
  booktitle={Proceedings of the European conference on computer vision (ECCV)},
  pages={384--400},
  year={2018}
}

@inproceedings{zareian2021open,
  title={Open-vocabulary object detection using captions},
  author={Zareian, Alireza and Rosa, Kevin Dela and Hu, Derek Hao and Chang, Shih-Fu},
  booktitle={Proceedings of the IEEE/CVF Conference on Computer Vision and Pattern Recognition},
  pages={14393--14402},
  year={2021}
}

@inproceedings{lin2014microsoft,
  title={Microsoft coco: Common objects in context},
  author={Lin, Tsung-Yi and Maire, Michael and Belongie, Serge and Hays, James and Perona, Pietro and Ramanan, Deva and Doll{\'a}r, Piotr and Zitnick, C Lawrence},
  booktitle={Computer Vision--ECCV 2014: 13th European Conference, Zurich, Switzerland, September 6-12, 2014, Proceedings, Part V 13},
  pages={740--755},
  year={2014},
  organization={Springer}
}

@inproceedings{gupta2019lvis,
  title={Lvis: A dataset for large vocabulary instance segmentation},
  author={Gupta, Agrim and Dollar, Piotr and Girshick, Ross},
  booktitle={Proceedings of the IEEE/CVF conference on computer vision and pattern recognition},
  pages={5356--5364},
  year={2019}
}

@article{zhou2019objects,
  title={Objects as points},
  author={Zhou, Xingyi and Wang, Dequan and Kr{\"a}henb{\"u}hl, Philipp},
  journal={arXiv preprint arXiv:1904.07850},
  year={2019}
}

@InProceedings{Steininger2021AircraftContextDataset,
    author    = {Steininger, Daniel and Widhalm, Verena and Simon, Julia and Kriegler, Andreas and Sulzbachner, Christoph},
    title     = {The Aircraft Context Dataset: Understanding and Optimizing Data Variability in Aerial Domains},
    booktitle = {Proceedings of the IEEE/CVF International Conference on Computer Vision (ICCV) Workshops},
    month     = {October},
    year      = {2021},
    pages     = {3823-3832}
}

@article{wang2024ov,
  title={Ov-vg: A benchmark for open-vocabulary visual grounding},
  author={Wang, Chunlei and Feng, Wenquan and Li, Xiangtai and Cheng, Guangliang and Lyu, Shuchang and Liu, Binghao and Chen, Lijiang and Zhao, Qi},
  journal={Neurocomputing},
  volume={591},
  pages={127738},
  year={2024},
  publisher={Elsevier}
}

@dataset{nakamura2020mad,
  author={T Nakamura},
  title={Military Aircraft Detection},
  year={2020},
  url={https://www.kaggle.com/datasets/a2015003713/militaryaircraftdetectiondataset}
}

@article{gu2021open,
  title={Open-vocabulary object detection via vision and language knowledge distillation},
  author={Gu, Xiuye and Lin, Tsung-Yi and Kuo, Weicheng and Cui, Yin},
  journal={arXiv preprint arXiv:2104.13921},
  year={2021}
}

@article{wu2024towards,
  title={Towards open vocabulary learning: A survey},
  author={Wu, Jianzong and Li, Xiangtai and Xu, Shilin and Yuan, Haobo and Ding, Henghui and Yang, Yibo and Li, Xia and Zhang, Jiangning and Tong, Yunhai and Jiang, Xudong and others},
  journal={IEEE Transactions on Pattern Analysis and Machine Intelligence},
  year={2024},
  publisher={IEEE}
}

@inproceedings{liu2019learning,
  title={Learning to assemble neural module tree networks for visual grounding},
  author={Liu, Daqing and Zhang, Hanwang and Wu, Feng and Zha, Zheng-Jun},
  booktitle={Proceedings of the IEEE/CVF International Conference on Computer Vision},
  pages={4673--4682},
  year={2019}
}

@article{wei2022rpc,
  title={RPC: a large-scale and fine-grained retail product checkout dataset},
  author={Wei, Xiu-Shen and Cui, Quan and Yang, Lei and Wang, Peng and Liu, Lingqiao and Yang, Jian},
  journal={Science China. Information Sciences},
  volume={65},
  number={9},
  pages={197101},
  year={2022},
  publisher={Springer Nature BV}
}

@inproceedings{plummer2017phrase,
  title={Phrase localization and visual relationship detection with comprehensive image-language cues},
  author={Plummer, Bryan A and Mallya, Arun and Cervantes, Christopher M and Hockenmaier, Julia and Lazebnik, Svetlana},
  booktitle={Proceedings of the IEEE international conference on computer vision},
  pages={1928--1937},
  year={2017}
}

@article{liu2023grounding,
  title={Grounding dino: Marrying dino with grounded pre-training for open-set object detection},
  author={Liu, Shilong and Zeng, Zhaoyang and Ren, Tianhe and Li, Feng and Zhang, Hao and Yang, Jie and Li, Chunyuan and Yang, Jianwei and Su, Hang and Zhu, Jun and others},
  journal={arXiv preprint arXiv:2303.05499},
  year={2023}
}

@inproceedings{zong2023detrs,
  title={Detrs with collaborative hybrid assignments training},
  author={Zong, Zhuofan and Song, Guanglu and Liu, Yu},
  booktitle={Proceedings of the IEEE/CVF international conference on computer vision},
  pages={6748--6758},
  year={2023}
}

@article{tian2020fcos,
  title={FCOS: A simple and strong anchor-free object detector},
  author={Tian, Zhi and Shen, Chunhua and Chen, Hao and He, Tong},
  journal={IEEE transactions on pattern analysis and machine intelligence},
  volume={44},
  number={4},
  pages={1922--1933},
  year={2020},
  publisher={IEEE}
}

@article{li2020generalized,
  title={Generalized focal loss: Learning qualified and distributed bounding boxes for dense object detection},
  author={Li, Xiang and Wang, Wenhai and Wu, Lijun and Chen, Shuo and Hu, Xiaolin and Li, Jun and Tang, Jinhui and Yang, Jian},
  journal={Advances in Neural Information Processing Systems},
  volume={33},
  pages={21002--21012},
  year={2020}
}

@inproceedings{kim2020probabilistic,
  title={Probabilistic anchor assignment with iou prediction for object detection},
  author={Kim, Kang and Lee, Hee Seok},
  booktitle={Computer Vision--ECCV 2020: 16th European Conference, Glasgow, UK, August 23--28, 2020, Proceedings, Part XXV 16},
  pages={355--371},
  year={2020},
  organization={Springer}
}

@inproceedings{zhou2022detecting,
  title={Detecting twenty-thousand classes using image-level supervision},
  author={Zhou, Xingyi and Girdhar, Rohit and Joulin, Armand and Kr{\"a}henb{\"u}hl, Philipp and Misra, Ishan},
  booktitle={European Conference on Computer Vision},
  pages={350--368},
  year={2022},
  organization={Springer}
}

@article{ren2016faster,
  title={Faster R-CNN: Towards real-time object detection with region proposal networks},
  author={Ren, Shaoqing and He, Kaiming and Girshick, Ross and Sun, Jian},
  journal={IEEE transactions on pattern analysis and machine intelligence},
  volume={39},
  number={6},
  pages={1137--1149},
  year={2016},
  publisher={IEEE}
}

@inproceedings{tan2021efficientnetv2,
  title={Efficientnetv2: Smaller models and faster training},
  author={Tan, Mingxing and Le, Quoc},
  booktitle={International conference on machine learning},
  pages={10096--10106},
  year={2021},
  organization={PMLR}
}

@article{xiao2024towards,
  title={Towards Visual Grounding: A Survey},
  author={Xiao, Linhui and Yang, Xiaoshan and Lan, Xiangyuan and Wang, Yaowei and Xu, Changsheng},
  journal={arXiv preprint arXiv:2412.20206},
  year={2024}
}

@inproceedings{bravo2023open,
  title={Open-vocabulary attribute detection},
  author={Bravo, Maria A and Mittal, Sudhanshu and Ging, Simon and Brox, Thomas},
  booktitle={Proceedings of the IEEE/CVF conference on computer vision and pattern recognition},
  pages={7041--7050},
  year={2023}
}

@inproceedings{pham2021learning,
  title={Learning to predict visual attributes in the wild},
  author={Pham, Khoi and Kafle, Kushal and Lin, Zhe and Ding, Zhihong and Cohen, Scott and Tran, Quan and Shrivastava, Abhinav},
  booktitle={Proceedings of the IEEE/CVF conference on computer vision and pattern recognition},
  pages={13018--13028},
  year={2021}
}

@article{torralba2010labelme,
  title={Labelme: Online image annotation and applications},
  author={Torralba, Antonio and Russell, Bryan C and Yuen, Jenny},
  journal={Proceedings of the IEEE},
  volume={98},
  number={8},
  pages={1467--1484},
  year={2010},
  publisher={IEEE}
}
\end{document}